\documentclass[sigconf]{acmart}
\usepackage{graphics}
\usepackage{amsmath}
\usepackage{amsfonts}
\usepackage{color}
\usepackage{multirow}
\usepackage{subfigure}
\usepackage{mathtools}
\usepackage{mathrsfs}
\usepackage{algorithm, algorithmic}

\graphicspath{{Figure/}}
\AtBeginDocument{%
  \providecommand\BibTeX{{%
    \normalfont B\kern-0.5em{\scshape i\kern-0.25em b}\kern-0.8em\TeX}}}

\copyrightyear{2021}
\acmYear{2021}
\setcopyright{acmcopyright}
\acmConference[KDD '21] {Proceedings of the 27th ACM SIGKDD Conference on Knowledge Discovery and Data Mining}{August 14--19, 2021}{Virtual Event, Singapore.}
\acmBooktitle{Proceedings of the 27th ACM SIGKDD Conference on Knowledge Discovery and Data Mining (KDD '21), August 14--19, 2021, Virtual Event, Singapore}
\acmPrice{15.00}
\acmISBN{978-1-4503-8332-5/21/08}
\acmDOI{10.1145/3447548.3467225}

\begin{document}
\fancyhead{}

\title{Deep Clustering based Fair Outlier Detection}


\author{Hanyu Song, Peizhao Li, Hongfu Liu}
\affiliation{%
  \institution{Michtom School of Computer Science, Brandeis University}
  \city{Waltham}
  \state{MA}
  \country{USA}
}
\email{{shydaniel,peizhaoli,hongfuliu}@brandeis.edu}

\renewcommand{\shortauthors}{Research Track Paper}

\begin{abstract}
In this paper, we focus on the fairness issues regarding unsupervised outlier detection. Traditional algorithms, without a specific design for algorithmic fairness, could implicitly encode and propagate statistical bias in data and raise societal concerns. To correct such unfairness and deliver a fair set of potential outlier candidates, we propose Deep Clustering based Fair Outlier Detection (DCFOD) that learns a good representation for utility maximization while enforcing the learnable representation to be subgroup-invariant on the sensitive attribute. Considering the coupled and reciprocal nature between clustering and outlier detection, we leverage deep clustering to discover the intrinsic cluster structure and out-of-structure instances. Meanwhile, an adversarial training erases the sensitive pattern for instances for fairness adaptation. Technically, we propose an instance-level weighted representation learning strategy to enhance the joint deep clustering and outlier detection, where the dynamic weight module re-emphasizes contributions of likely-inliers while mitigating the negative impact from outliers. Demonstrated by experiments on eight datasets comparing to 17 outlier detection algorithms, our DCFOD method consistently achieves superior performance on both the outlier detection validity and two types of fairness notions in outlier detection. 
\end{abstract}



\maketitle
\section{Introduction}

Utility-oriented machine learning systems have trickled down to the real-world in aid of high-stake decision making in various fields. The inherent bias brought by unbalanced data and embedded in those models, if not controlled with specific algorithmic design, will inevitably backfire and deteriorate the existing social barricades. Due to the rising societal concerns, fairness in machine learning has received increasing attention in recent years~\cite{chouldechova2018frontiers,hashimoto2018fairness,kearns2019empirical,li2020deep}. Outlier detection is one of the most vulnerable domains in face of the lurking algorithmic unfairness. The task is to find rare or suspicious individuals that deviate from the majority. Due to the strong correlation between the minority and outliers, an outlier detector might easily suffer from discrimination against certain sensitive attributes like ethnicity or gender. For instance, in credit risk assessment, credit card applicants may suffer from a biased credit scoring algorithm against certain sexuality~\cite{creditcard}; racial equality will deteriorate when criminality detection heavily hinges on the object's appearance~\cite{facecrime}. To recognize the significance of equality preservation, we need to address the urgent demands of fairness-aware outlier detection methods. However, this recently-emerged task has received little attention before 2020 and has not been fully addressed yet.

On a colloquial formulation, unsupervised fair outlier detection aims to find potential outlier candidates that substantially differ from the majority instances while maintaining insignificant to sensitive attribute subgroups\footnote{In the paper we use "sensitive attribute subgroup", "sensitive subgroup", and "protected subgroup" interchangeably.} (i.e., gender or ethnicity)~\cite{davidson2020framework}. To our best knowledge, only two methods exist along this direction and are both proposed in 2020. \citet{FLOF} propose the FairLOF algorithm, which improves classical LOF~\cite{LOF} in regards to three heuristic principles: neighborhood diversity, apriori distribution, and attribute asymmetry, to prevent an unfair sensitive subgroup distribution among top-ranked outlier candidates. \citet{shekhar2020fairod} design FairOD, a deep model using a standard autoencoder component for self-reconstruction with the statistical parity and group fidelity as fairness constraints to ensure the majority protected subgroup share the same outlier rate as the minority one.

Although the above attempts have made pioneering progress in addressing this newly emerging yet crucial problem, we see some limitations in the existing work that could be further improved. FairLOF operates on the original feature space, which limits its capacity towards detection accuracy and fairness degree. Since the original feature space often does not expose much information on the underlying data pattern, it adds difficulty to separate anomalies from inliers. FairOD employs the classic autoencoder for representation learning via minimizing the self-reconstruction loss and uses it as the outlier detection criteria. However, such a deep embedding treats all samples indifferently. It does not filter the noise and disruptions that outliers may impose on the feature space, thus preventing itself from pursuing high performance. Moreover, existing methods evaluate fairness degree only in terms of the outlier rates in different groups, while ignoring the disparate predictive validity between groups, which could also lead to unjust results. Consider the case where an outlier detector returns a consistent $5\%$ outlier rate for two sensitive subgroups, but the true positive rates for two groups are $5\%$ and $10\%$, respectively. The detector yields a huge performance penalty on the second subgroup, yet such a performance gap is unrecognizable under the existing evaluative phase. Finally, existing methods mostly perform evaluations on a limited number of benchmarks, including synthetic datasets. An extensive exploration with more real-world datasets is heavily needed.

To solve the above challenges, we propose a novel Deep Clustering based Fair Outlier Detection (DCFOD) method. Inspired by the reciprocal relationship between outlier detection and clustering~\cite{chawla2013k,COR}, we employ a joint deep clustering/outlier detection framework and propose weighted representation learning towards each instance. The degree of outlierness for each individual is associated with the distance to its nearest cluster centroid. The dynamic weight module enhances outlier detection validity by re-emphasizing likely-inliers’ contributions while mitigating outliers’ negative impacts. To simultaneously ensure fairness adaptation, we utilize fairness-adversarial training with a min-max strategy to conceal sensitive information and achieve group fairness, while preserving the feature validity on the embedded representation. In terms of fairness evaluation, we propose two metrics to measure the degree of fairness, based on the diagnostic ability gap among subgroups and the subgroup distribution drift among top-ranked outliers. To sum up, we underline our contributions as follows:

\begin{itemize}
    \item We address the newly-emerged fair outlier detection problem and propose a novel Deep Clustering based Fair Outlier Detection (DCFOD) framework.
    \item Our DCFOD adopts representation learning and fairness-adversarial training, with a novel dynamic weight in regulation of negative impacts from outlier points, to obtain a downstream task-favorable representation while simultaneously ensuring improvement in fairness degree. 
    \item We strengthen the fairness measurements in the context of outlier detection by proposing two fairness metrics that test subgroup-wise agnostic ability gap and subgroup distribution drift in detected outliers. 
    \item Extensive experiments on eight real-world public datasets demonstrate evident edge of DFCOD on all metrics in competitions with 17 unsupervised outlier detection methods, including recently-proposed fair outlier detection methods and other conventional outlier detection algorithms.
\end{itemize}
\vspace{-4mm}


\section{Related Work}

In this section, we illustrate the related work in terms of unsupervised outlier detection, deep outlier detection, fair machine learning, and recent advances in fair outlier detection.

\noindent{\textbf{Unsupervised Outlier Detection}}. Typical unsupervised outlier detection algorithms calculate a continuous score for each data point to quantify its outlier degree. Based on diverse assumptions, a number of outlier detection methods have been proposed, including linear models~\cite{PCA,OCSVM}, proximity-based models~\cite{LOF,COF,CBLOF,HBOS}, and probability-based models~\cite{FABOD,COPOD}. Moreover, some studies purse outlier detection by subspace learning~{\cite{10.1007/978-3-642-01307-2_86}}, low-rank~{\cite{zhao2015dual}}, matrix-completion~{\cite{kannan2017outlier}}, and random walk~\cite{pang2016outlier}. Since most above-mentioned methods rely heavily on various assumptions, ensemble-based outlier detectors are brought up to alleviate assumption dependence, including iForest~{\cite{iforest}}, bi-sampling outlier detection~{\cite{liu2016outlier}}, feature bagging~{\cite{FB}}, lightweight online detector of anomalies~{\cite{LODA}}, and clustering with outlier removal~\cite{COR}. 

\noindent{\textbf{Deep Outlier Detection}}. Recent advances in representation learning~\cite{he2016deep,krizhevsky2017imagenet} have demonstrated that deep neural networks are capable of extracting effective features for downstream tasks. The key in deep outlier detection is to seek a self-supervised signal for representation learning. Autoencoder~\cite{AE} and variational autoencoder~\cite{VAE} are two widely used frameworks that optimize the self-reconstruction loss. Their variants include robust autoencoder with $L_1$ sparse constraint~\cite{zhou2017anomaly}, autoencoder ensemble~\cite{chen2017outlier} and stacked autoencoder~\cite{wan2019outlier}. Other state-of-the-art deep approaches include random mapping~\cite{ijcai2020-408}, which predicts proximity information in a randomly projected space; generative adversarial networks~{\cite{zenati2018efficient,li2018anomaly}}, which focuses on generating fake outliers by min-max training; predictability models~\cite{liu2018future}, which predicts the current data instances using the representations of the previous instances within a temporal window. More details on deep outlier detection can be found in this recent review~\cite{pang2020deep}.

\begin{figure*}[t]
	\centering
	\includegraphics[width=0.85\textwidth]{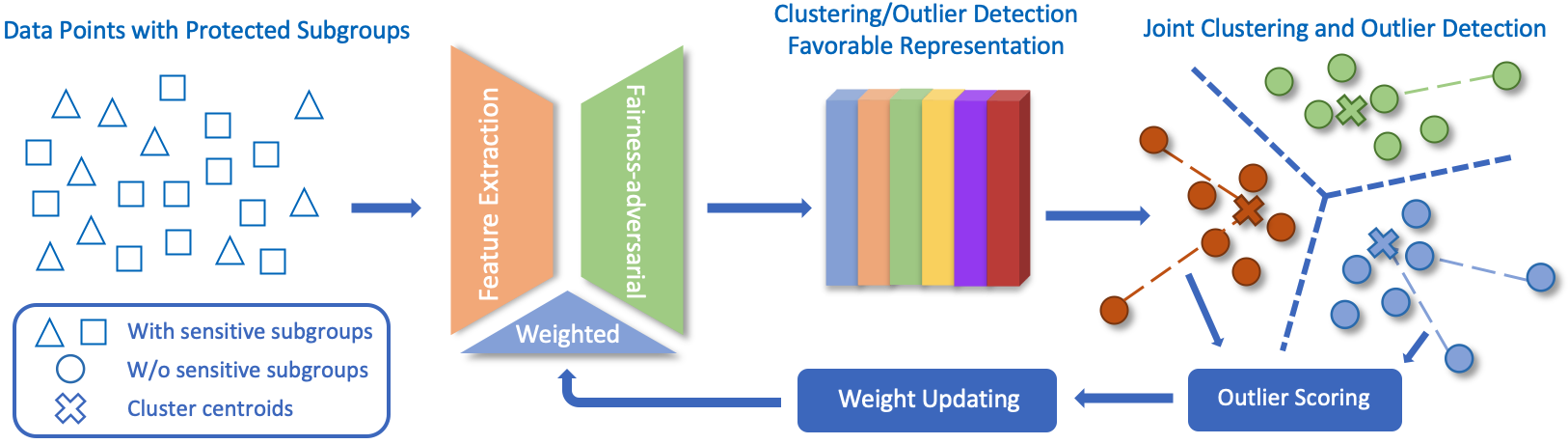}\vspace{-3mm}
	\caption{Overview of Deep Clustering based Fair Outlier Detection. The model conducts weighted representation learning with min-max training between the feature extractor and fairness-adversarial subgroup discriminator to conceal the protected subgroups on the learned representation and eventually achieve subgroup-invariant outlier detection. Distance-based outlier scoring then propagates weight adjustment to guide learning on the next iteration.}\label{fig:framework}\vspace{-3mm}
\end{figure*}

\noindent{\textbf{Fair Machine Learning and Fair Outlier Detection}}. According to various application scenarios, abundant fairness notions have been proposed, including individual fairness~\cite{dwork2012fairness}, where similar individuals are expected to be treated in a similar fashion; group fairness~\cite{li2021dyadic,dwork2012fairness,kleinberg2016inherent}, where samples in different groups ought to be treated equally; subgroup fairness~\cite{kearns2018preventing}, which extends group fairness by demanding fairness on combinations of groups (i.e. subgroups); causality-based fairness~\cite{kusner2017counterfactual, chiappa2019path}, where causal inference on multiple features is conducted to track the relationship between sensitive attributes and samples while mitigating historical bias. Aside from fairness definitions, debiasing techniques can be categorized into three main approaches: pre-processing~\cite{feldman2015certifying}, in-processing~\cite{zemel2013learning, madras2018learning} and post-processing~\cite{postprocess}. The deep representation learning is a typical in-processing method that incorporates fairness constraints into the model optimization objectives to obtain an ideal model parameterization that maximizes performance and fairness~\cite{caton2020fairness}. In particular, representation learning with adversary has become a widely-used method in recent years and has demonstrated effectiveness on multiple tasks, including anonymization~\cite{edwards2015censoring}, clustering~\cite{li2020deep}, classification~\cite{madras2018learning}, transfer learning~\cite{madras2018learning}, and domain adaptation~\cite{tzeng2017adversarial,li2020mining}.

Existing two algorithms in pursuit of fair outlier detection typically target group fairness and contain a post-processing technique that seeks an outlier ranking that is consistent with the baseline result. FairLOF~\cite{FLOF} is the first paper to address the fair outlier detection problem, which incorporates a corrective term on the baseline LOF algorithm, in regards to local sensitive subgroup diversity and global outlier alignment with the baseline. FairOD~\cite{shekhar2020fairod} is another pioneering work, which has recently been released on arXiv. This method targets an equal outlier rate on the majority and minority sensitive subgroups. Specifically, it employs an autoencoder~\cite{AE} as the base outlier detector and performs subgroup debiasing with statistical parity fairness constraint, while maintaining fidelity to within-group rankings with respect to the baseline.

However, one concern the post-processing approach may have is the baseline restricting the fair model's detection validity, due to their aligned outlier ranking. Therefore, we set to tackle the fair outlier detection problem with an in-processing deep representation learning methodology, where we ensure detecting validity on the learned task-friendly feature space while conducting fairness-adversarial training to concurrently maintain group fairness.


\section{Fair Outlier Detection}

Given a set of data instances with categorical sensitive attributes, unsupervised fair outlier detection aims to seek top outlier candidates in a sensitive attribute-invariant fashion. Specifically, we define the unsupervised fair outlier detection problem as follows:

\noindent\textbf{Problem Formulation} (Unsupervised fair outlier detection). \textit{Given data points $X = \{x_i\}_{i = 1}^N$ in space $\mathcal{X}$ sampled i.i.d. from an underlying distribution $\mathcal{D}$, with sensitive attribute $S = \{s_i\}_{i = 1}^N$ ranging in $M$ subgroup classes $C = \{c_i\}_{i = 1}^M$ and inaccessible ground truth outlier label $Y$, the problem of unsupervised fair outlier detection aims to seek an outlier score assignment $O = \{o_i\}_{i = 1}^N$ that faithfully distinguishes the underlying outliers from the majority inliers, while maintaining statistical independence to sensitive subgroups,}
\begin{equation}~\label{eq:def1}
    \Pr[o_i = o\vert s_i = c] = \Pr[o_i = o], \forall o \in \mathbb{R}, c \in C, i \in [1, N],
\end{equation}
\textit{and holding an accuracy parity between sensitive subgroups,}
\begin{equation}~\label{eq:def2}
    \mathbb{E}_{\mathcal{D}_i}[\left| Y - O\right|\vert S = c_i] = \mathbb{E}_{\mathcal{D}_j}[\left| Y - O\right|\vert S = c_j], \forall i, j \in [1,M],
\end{equation}

Eq.~\eqref{eq:def1} and ~\eqref{eq:def2} both reflect the targeted group fairness~\cite{dwork2012fairness,kleinberg2016inherent}. In Eq.~\eqref{eq:def1}, the probability distribution of $O$ is invariant to sensitive subgroups; In Eq.~\eqref{eq:def2}, where $\mathcal{D}_i$ denotes the underlying data distribution for subgroup $i$, the detection discrepancy between the true labels and outlier scores ought to be the same for all subgroups.

Existing methods~\cite{FLOF,shekhar2020fairod} on fair outlier detection achieve group fairness by balancing out the protected subgroup diversity, which adds difficulty to outlier uncovering as it blurs the underlying data pattern. To alleviate that downside, they preserve a consistent outlier ranking with respect to a pre-obtained baseline result, yet this may also impose restrictions on detection validity. Motivated by the above limitations, we adopt the representation learning with adversarial training framework, which guarantees the detection accuracy on a task-favorable feature space, while achieving the expected group fairness by hiding sensitive subgroup information on the learned representation.

\section{Methodology}

We elaborate our proposed Deep Clustering based Fair Outlier Detection (DCFOD), using symbols summarized in Table~\ref{tab:notation}.

\begin{table}[!t]
\small
\caption{Main notations and descriptions}
\vspace{-4mm}
\centering
\begin{tabular}{c|c| l} \hline
    Notation & Type & Description \\ \hline 
    $N$ & Input &Number of instances \\
    $M$ & Input &Number of subgroup classes \\
    $D$ & Input &Dimension of embedded feature space \\
    $K$ & Input& Number of clusters \\
    $x_i$ & Input& Vector of $i$-th data point \\
    $s_i$ & Input& Sensitive subgroup of $i$-th data point \\
    \hline
    $c_i$ & Learnable & Cluster membership of $i$-th data point \\
    $o_i$ & Learnable & Outlier score of $i$-th data point \\
    $w_i$ & Learnable & Weight of $i$-th data point \\
    $\mu_k$ & Learnable & Vector of $k$-th centroid \\
\hline
\end{tabular}\vspace{-4mm}
\label{tab:notation}
\end{table}
\subsection{Framework Overview}

Unsupervised fair outlier detection can be decomposed into unsupervised outlier detection and fair learning. Inspired by the huge progress in deep neural networks, we tackle the above two problems under a same umbrella with representation learning. Figure~\ref{fig:framework} shows our proposed DCFOD framework. Administered by a dynamic weight, the representation learning of DCFOD consists of a feature extractor and a fairness-adversarial counterpart. For fairness-adaptation, we conduct a min-max training, where a sensitive subgroup discriminator strives to predict instances' subgroup category, while the feature extractor undermining that prediction. Expected group fairness is achieved when the prediction ceases to improve, due to the neutralization of the subgroup information on the learned feature embedding. The dynamic weight module secures the outlier detection validity amidst fair learning. Specifically, we perform a joint clustering and outlier detection on the embedded space. Given the strong coupling between clustering and outlier detection, the well-discovered cluster structure fosters outlier uncovering, while assisting the weight calculation. Data points close to their associated centroids are assigned higher weights while those far away from centroids, identified as potential outlier candidates, are assigned lower weights. Therefore, inlier points with dominant weights capture the intrinsic cluster structure, which, in turn, brings benefits to outlier detection. By this means, we diminish outliers' disturbance on representation learning, thus guaranteeing the outlier diagnostic quality on the obtained representation.


\subsection{Weighted Representation Learning}

We denote the feature extractor and subgroup discriminator as $f$ and $g$, where $f$ seeks a latent representation for samples $X$ and $g$ predicts its sensitive subgroup membership, i.e., $f: \mathcal{X}~\rightarrow \mathcal{Z} \subseteq \mathbb{R}^{D}$, $g: \mathcal{Z}~\rightarrow \mathbb{R}^{M}$. Furthermore, we formulate $K$ cluster centroids on the latent space: $\mu \subseteq \mathbb{R}^{K \times D}$, and calculate outlier score $o_i$ with dynamic weight $w_i$ for each instance $x_i$. Based on the above components, we establish the training objective as three loss functions: self-reconstruction loss $\mathcal{L}_s$, fairness-adversarial loss $\mathcal{L}_f$, and clustering regularization $\mathcal{L}_r$. In what follows, we elaborate the formulations of each component and the calculation of $o_i$ and $w_i$.

\noindent{\textbf{Self-reconstruction Loss}}. We employ a weighted autoencoder as the feature extractor, which contains an encoder $f$ that seeks a feature representation and a decoder $h: \mathcal{Z}~\rightarrow \mathcal{X}$ to reconstruct back to the original space. The weight module $w_i$, which will be elaborated in detail in Eq.~\eqref{eq: weight}, mitigates the negative impact from outliers’ abnormal feature representations on model parameter training. $\mathcal{L}_s$, with a L2 distance, is expressed as:
\begin{equation}
    \mathcal{L}_s \coloneqq \sum_i^N w_i \cdot \Vert x_i - h\circ f(x_i)\Vert_2^2,
\end{equation}
where $\circ$ denotes the function composition.

\noindent{\textbf{Fairness-adversarial Loss}}. Furthermore, the learned representation is also expected to be sensitive subgroup-invariant. To achieve this, we employ a sensitive attribute subgroup discriminator $g$ predicting each data point's likelihood of $M$ subgroup categories in the latent space. $\mathcal{L}_f$ records the predicting loss as follows:
\begin{equation}
\mathcal{L}_f \coloneqq \sum_i^N w_i \cdot \ell(g \circ f(x_i), \ s_i),
\end{equation}
where $\ell$ is the cross-entropy loss and $s_i$ is the true subgroup class for $x_i$. During the fairness min-max training, $g$ and $f$ are trained in opposite directions. Specifically, $g$ aims to accurately predict protected subgroups via minimizing $\mathcal{L}_f$, while $f$, in adversary, maximizes the loss by hiding protected information on the feature embedding. As training progresses, the obtained feature representation becomes subgroup-invariant for fair outlier detection.

\noindent{\textbf{Clustering Assignment and Regularization}}. To perform joint clustering and outlier detection, we first obtain a soft clustering assignment $P$ on the feature embedding. Specifically, we calculate the probability of assigning each $f(x_i) \in \mathcal{Z}$ to each of the $K$ clusters based on its distance to the centroids with Student's t-distribution:
\begin{equation}~\label{eq: p}
p_{ik} \coloneqq \frac{(1 + \Vert f(x_i) - \mu_k\Vert_2^2)^{-1}}{\sum_j^K (1 + \Vert f(x_i) - \mu_{j}\Vert_2^2)^{-1}},
\end{equation}
where $p_{ik}$ is the probability of assigning the $i$-th data point to the $k$-th cluster, and $\mu_k$ denotes the $k$-th cluster centroid. The degree of freedom for the t-distribution is 1. 

To strengthen the cluster predictive confidence, we follow~\citet{DEC} and employ a clustering regularizer, which computes an auxiliary distribution $Q$ based on the soft clustering assignment $P$ as follows:
\begin{equation}~\label{eq: q}
q_{ik} \coloneqq \frac{p_{ik}^2/\sum_i^N p_{ik}}{\sum_{j}^K p_{ij}^2/\sum_i^N p_{ij}}.
\end{equation}
In this way, the auxiliary distribution enlarges the distinctions between predictions on different clusters. The clustering regularizer then pulls $P$ toward $Q$ by minimizing their KL divergence. Similar to the above losses, $\mathcal{L}_r$ is also regulated by the weight module:
\begin{equation}
\mathcal{L}_r \coloneqq \sum_i^N w_i \cdot \textup{D}_{\textup{KL}}(P \Vert Q) = \sum_i^N \sum_k^K w_i \cdot p_{ik} \cdot \log\frac{p_{ik}}{q_{ik}}.
\end{equation}
\begin{algorithm}[!t]
\caption{DCFOD}
\label{alg: pseudo}
\small
\begin{algorithmic}[1]
\REQUIRE Data point $x_i$ with subgroup membership $s_i$, $i=1$ to $N$;\\ 
\ \ \ \ \ \  Cluster number $K$. Learning rate $\eta$.
\ENSURE Continuous outlier score $o_i$, $i=1$ to $N$.
\STATE Initialize parameters for $f, g$, and $h$;
\STATE $\mu \leftarrow Kmeans([f(x_1),\cdots,f(x_N)], K)$;
\WHILE{model not converged}
\STATE Calculate the soft assignment $p_{ik}$ and the auxiliary distribution $q_{ik}$ by Eq.~\eqref{eq: p} and \eqref{eq: q};
\STATE Calculate the outlier score of each data point $o_i$ by Eq.~\eqref{eq: os};
\STATE Calculate the weight of each data point $w_i$ by Eq.~\eqref{eq: weight};
\STATE Calculate $\bigtriangledown\mathcal{L}_f$ with $w_i, s_i$, $f$, and $g$;
\STATE Calculate $\bigtriangledown(\mathcal{L}_s + \mathcal{L}_r - \mathcal{L}_f)$ with $w_i$, $p_{ik}$, $q_{ik}$, $s_i$, $f$, $g$, $h$ and $\mu$; 
\STATE Update $g \leftarrow g - \eta\bigtriangledown\mathcal{L}_f$;
\STATE Update $(f, h, \mu) \leftarrow (f, h, \mu) - \eta\bigtriangledown(\mathcal{L}_s + \mathcal{L}_r - \mathcal{L}_f)$;

\ENDWHILE
\RETURN Outlier score $o_i$ for $i=1$ to $N$.
\end{algorithmic}
\end{algorithm}

\noindent{\textbf{Outlier Score}}. Outliers are conceptually defined by clusters as data points that deviate from the majority cluster structure. Based on that, we define outlierness as the within-cluster distance ratio to the centroid. Specifically, $o_i$ is computed by the Euclidean distance between $f(x_i)$ and its closest centroid, normalized by that of the furthest point belonging to the same centroid:
\begin{equation}~\label{eq: os}
o_i \coloneqq \frac{\min_{k\in [1, K]} \Vert f(x_i) - \mu_{k}\Vert_2}{\max_{j \in [1,N]} \mathbb{I}_{m_j = k}\cdot\Vert f(x_j) - \mu_{k}\Vert_2},
\end{equation}
where $\mathbb{I}$ is an indicator function, and $m_j$ is the most probable cluster membership for the $j$-th point, based on clustering distribution $P$. 

\noindent{\textbf{Weight Module}}. To mitigate negative impacts from outliers, we guide the representation learning with an instance-level weight module $w_i$, where we deemphasize outliers with lower weights compared to inliers. Specifically, $w_i$ in $\mathcal{L}_s$ alleviates the disruption in feature extractor's parameter updating due to outliers' abnormal feature representations; $w_i$ in $\mathcal{L}_r$ prevents strengthening the predictive confidence for outliers, as conceptually, they do not belong to any cluster; For $w_i$ in $\mathcal{L}_f$, the idea is to increase focus in subgroup debiasing on potential outlier candidates.

Therefore, with the above $o_i$, we compute the dynamic weight mechanism that governs the designated components in our objective as follows:
\begin{equation}~\label{eq: weight}
w_i \coloneqq \frac{e^{-o_i}}{\sum_j^N e^{-o_j}},
\end{equation}
which is essentially a \textit{Softmax} function imposed on the negative values of outlier scores. We design $w_i$ such that it is in negative correlation with $o_i$, while keeping $\sum_i^N w_i = 1$, a global balance that avoids weights excessively overvalued.

In conclusion, the overall objective function of our framework can be written as the following minimax optimization:
\begin{equation}
\begin{split}
\max_{f,\ h,\ \mu} \quad \beta\mathcal{L}_f - \alpha\mathcal{L}_s - \mathcal{L}_r,\quad \min_{g} \quad \mathcal{L}_f, 
\end{split}
\end{equation}
where $\alpha$ and $\beta$ are hyperparameters that control the balance in the objective function. In the end, our method returns a set of outlier scores that maintains detection validity and group fairness. The complete process of DCFOD is shown in Algorithm~\ref{alg: pseudo}.


\section{Experiment}

In this section, we first introduce the experimental settings, datasets and evaluative metrics for outlier validation and degree of fairness. Then we compete our model with 17 unsupervised outlier detection methods on eight public datasets. Finally, we conduct a factor exploration on our model for an in-depth analysis.

\subsection{Experimental Setup}

\noindent{\textbf{Datasets}}. We choose eight UCI~\footnote{https://archive.ics.uci.edu/ml/datasets.php} datasets in diverse fields, including credit rating, personal income, academic performance, drug review, and clinical screening. Table~\ref{tab:datasets} shows detailed characteristics of these datasets. All datasets except \textit{german} contain either \textit{Gender} or \textit{Race} as the sensitive attribute, while that of \textit{german} is provided as a combination of gender and marital status. The \textit{Outlier Definition} column shows the benchmark outlier criteria for each dataset, which is used as the ground truth for outlier validation. The numbers of instances in these datasets vary from 704 to nearly 300,000, and the outlier percentage ranges from 6.2\% to 30\%.

\noindent{\textbf{Competitive Methods}}. We compare our model with the fair outlier detectors FairLOF~\cite{FLOF}, FairOD~\cite{shekhar2020fairod}, and 15 conventional unsupervised outlier detection methods, including linear models: Principal Component Analysis (PCA)~\cite{PCA}, One-class Support Vector Machine (OCSVM)~\cite{OCSVM}; proximity-based models: Local Outlier Factor (LOF)~\cite{LOF}, Connectivity-Based Outlier Factor (COF)~\cite{COF}, Clustering Based Local Outlier Factor (CBLOF)~\cite{CBLOF}; probability-based models: Fast angle-based Outlier Detector (FABOD)~\cite{FABOD}, Copula Based Outlier Detector (COPOD)~\cite{COPOD}; ensemble-based models: Feature Bagging (FB)~\cite{FB}, iForest~{\cite{iforest}}, Lightweight On-line Detector of Anomalies (LODA)~\cite{LODA}, Clustering with Outlier Removal (COR)~\cite{COR}; neural networks: AutoEncoder (AE)~\cite{AE}, Variational Auto Encoder (VAE)~\cite{VAE}, Random Distance Prediction (RDP)~\cite{ijcai2020-408}.

\noindent{\textbf{Implementation}}. We implement FairLOF, FairOD in PyTorch, build RDP, COR based on their public source code, and other classical outlier detection methods are implemented with PyOD~\cite{pyod}. In practice, we use the following default settings. The hyperparameters for FairLOF and FairOD strictly follow their literature. The number of nearest neighbors for LOF, COF, and FABOD are set to 20; the number of clusters for CBLOF and COR are set to 10; the number of bins and random cuts for LODA is 10 and 100, respectively; the number of base estimators and sub-sampling size for iForest are 100 and 256 respectively; the base estimator for feature bagging is LOF, its number of estimators is 10 and its sub-sampling size equals to the sample size. For AE and VAE, the hidden neuron structures are 64-32-32-64 and 128-64-32-32-64-128. In cases when the feature dimension is smaller than 64, we modify their structures into 16-8-8-16 and 16-8-4-4-8-16, accordingly. Both of them run for 100 epochs with a batch size of 32. 
\begin{table}[t]
    \caption{Characteristics of datasets}\vspace{-4mm}
    \label{tab:datasets}
    \centering
    \resizebox{.475\textwidth}{!}{%
        \begin{tabular}{c | cccccc}
            \toprule
            Dataset & \#Instances & \#Features & Domain & Sensitive Attribute & Outlier Definition & \%Outlier\\
            \midrule
            \it{student} & 1045 & 33 & Academic & Gender & Final grade $\leq$ 6 & 9.58\%\\
            \it{asd} & 704 & 21 & Clinical & Gender & Should pursue diagnosis &26.84\%\\
            \it{obesity} & 2111 & 17 & Clinical & Gender & Insufficient weight &13.60\%\\
            \it{cc} & 30000 & 24 & Credit & Gender & Credit default & 22.12\% \\
            \it{german} & 1000 & 21 & Credit & Marital \& Gender & Good credit &30.00\%\\
            \it{drug} & 1885 & 13 & Drug & Gender & Used within last week & 8.44\%\\
            \it{adult} & 48842 & 14 & Income & Race & Income > 50K & 23.93\%\\
            \it{kdd} & 299285 & 40 & Income & Race & Income > 50K & 6.2\%\\
            \bottomrule        
        \end{tabular}}\vspace{-4mm}
\end{table}

\begin{table*}[t]
    \caption{Outlier validation comparison of 18 algorithms on 8 benchmark datasets by $AUC$}\vspace{-4mm}
    \label{tab:auc}
    \centering
    \resizebox{\textwidth}{!}{%
        \begin{tabular}{c|cc|cc|ccc|cc|cccc|ccc|cc}
            \toprule
            \multirow{2}{*}{Dataset} 
                & \multicolumn{2}{c|}{Fair Model} &
                \multicolumn{2}{c|}{Linear Model} & \multicolumn{3}{c|}{Proximity Based} & \multicolumn{2}{c|}{Probability Based} & \multicolumn{4}{c|}{Ensemble Based} & \multicolumn{3}{c|}{Neural Networks} & \multicolumn{2}{c}{Ours}\\ \cmidrule{2 - 19}
            {} & FairLOF & FairOD & PCA & OCSVM & LOF & COF & CBLOF & FABOD & COPOD & FB & iForest & LODA & COR & AE & VAE & RDP  & \textbf{DCOD} & \textbf{DCFOD}\\
            \midrule
            \it{student} & 0.662 & 0.717\scriptsize{$\pm$0.123} & 0.718 & 0.848 & 0.777 & 0.710 & 0.761 & 0.834 & 0.750 & 0.804 & 0.679 & 0.794\scriptsize{$\pm$0.054} & 0.568 & 0.728 & 0.719 & 0.834\scriptsize{$\pm$0.021} & {\it\color{blue}0.912}\scriptsize{$\pm$0.011} & {\bf\color{red}0.921}\scriptsize{$\pm$0.013}					\\
            
            \it{asd} & 0.543 & 0.642\scriptsize{$\pm$0.214} & 0.630 & 0.707 & 0.523 & 0.491 & 0.387 & 0.469 & 0.811 & 0.521 & 0.542 & 0.524\scriptsize{$\pm$0.152} & 0.857 & 0.627 & 0.629 & 0.695\scriptsize{$\pm$0.037} & {\it\color{blue}0.939}\scriptsize{$\pm$0.038} & {\bf\color{red}0.945}\scriptsize{$\pm$0.032}\\
            
            \it{obesity} & 0.428 & 0.514\scriptsize{$\pm$0.167} & 0.720 & 0.675 & 0.355 & 0.377 & 0.602 & 0.540 & 0.742 & 0.383 & 0.757 & 0.572\scriptsize{$\pm$0.134} & 0.552 & 0.719 & 0.719 & 0.612\scriptsize{$\pm$0.021} & {\it\color{blue}0.800}\scriptsize{$\pm$0.029} & {\bf\color{red}0.825}\scriptsize{$\pm$0.017}					\\
            
            \it{cc} & 0.465 & 0.554\scriptsize{$\pm$0.016} & 0.604 & {\bf\color{red}0.621} & 0.451 & 0.465 & 0.579 & 0.531 & 0.557 & 0.451 & 0.594 & {\it\color{blue}0.605}\scriptsize{$\pm$0.001} & 0.546 & 0.604 & 0.604 & 0.597\scriptsize{$\pm$0.011} & 0.571\scriptsize{$\pm$0.018} & 0.566\scriptsize{$\pm$0.024}					\\
            
            \it{german} & 0.569 & 0.570\scriptsize{$\pm$0.021} & 0.546 & 0.584 & {\bf\color{red}0.589} & 0.568 & 0.581 & {\bf\color{red}0.589} & 0.547 & {\it\color{blue}0.582} & 0.546 & 0.506\scriptsize{$\pm$0.034} & 0.475 & 0.546 & 0.546 & 0.581\scriptsize{$\pm$0.030} & 0.558\scriptsize{$\pm$0.016} & 0.559\scriptsize{$\pm$0.013}					\\
            
            \it{drug} & 0.543 & 0.551\scriptsize{$\pm$0.035} & 0.562 & 0.562 & 0.546 & 0.534 & 0.555 & 0.562 & 0.556 & 0.544 & 0.570 & 0.521\scriptsize{$\pm$0.055} & 0.522 & 0.560 & 0.562 & 0.567\scriptsize{$\pm$0.013} & {\bf\color{red}0.588}\scriptsize{$\pm$0.031} & {\it\color{blue}0.586}\scriptsize{$\pm$0.021}					\\
            
            \it{adult} & 0.486 & {\bf\color{red}0.605}\scriptsize{$\pm$0.022} & 0.473 & 0.525 & 0.488 & 0.479 & 0.474 & 0.471 & 0.488 & 0.438 & 0.426 & 0.504\scriptsize{$\pm$0.095} & 0.442 & 0.473 & 0.473 & 0.541\scriptsize{$\pm$0.014} & 0.487\scriptsize{$\pm$0.072} & {\it\color{blue}0.592}\scriptsize{$\pm$0.051}					\\
            
            \it{kdd} & N/A$^*$ & 0.706\scriptsize{$\pm$0.003} & 0.661 & 0.630 & 0.549 & N/A & 0.687 & 0.519 & 0.686 & 0.566 & 0.589 & 0.472\scriptsize{$\pm$0.114} & 0.424 & 0.661 & 0.661 & 0.675\scriptsize{$\pm$0.000} & {\bf\color{red}0.752}\scriptsize{$\pm$0.007} & {\it\color{blue}0.747}\scriptsize{$\pm$0.011}					\\
            \midrule
            $Score_{AUC}$ & 0.751 & 0.852 & 0.854 & 0.892 & 0.752 & 0.737 & 0.816 & 0.792 & 0.882 & 0.751 & 0.786 & 0.751 & 0.821 & 0.855 & 0.854 & 0.886 & {\it\color{blue}0.953} & {\bf\color{red}0.979}					\\
            \bottomrule
            \multicolumn{19}{l}{$^*$N/A means out-of-memory error. For this method, the according dataset is not included in the $Score$ measurement calculation.}
        \end{tabular}}\vspace{-2mm}
\end{table*}

\begin{table*}[t]
    \caption{Degree of fairness comparison of 18 algorithms on 8 benchmarks by $F_{Gap}$}\vspace{-4mm}
    \label{tab:gap}
    \centering
    \resizebox{\textwidth}{!}{%
        \begin{tabular}{c|cc|cc|ccc|cc|cccc|ccc|cc}
            \toprule
            \multirow{2}{*}{Dataset} 
                & \multicolumn{2}{c|}{Fair Model} &
                \multicolumn{2}{c|}{Linear Model} & \multicolumn{3}{c|}{Proximity Based} & \multicolumn{2}{c|}{Probability Based} & \multicolumn{4}{c|}{Ensemble Based} & \multicolumn{3}{c|}{Neural Networks} & \multicolumn{2}{c}{Ours}\\ \cmidrule{2 - 19}
            {} & FairLOF & FairOD & PCA & OCSVM & LOF & COF & CBLOF & FABOD & COPOD & FB & iForest & LODA & COR & AE & VAE & RDP  & \textbf{DCOD} & \textbf{DCFOD}\\
            \midrule
            \it{student} & 0.008 & 0.043\scriptsize{$\pm$0.027} & 0.053 & 0.042 & {\bf\color{red}0.001} & 0.069 & 0.100 & 0.011 & 0.049 & 0.009 & 0.073 & 0.060\scriptsize{$\pm$0.040} & {\it\color{blue}0.003} & 0.049 & 0.053 & 0.028\scriptsize{$\pm$0.006} & 0.020\scriptsize{$\pm$0.013} & 0.012\scriptsize{$\pm$0.011}					\\
            
            \it{asd} & {\it\color{blue}0.013} & 0.024\scriptsize{$\pm$0.019} & 0.034 & 0.040 & 0.032 & 0.016 & 0.068 & {\bf\color{red}0.004} & 0.016 & {\bf\color{red}0.004} & {\it\color{blue}0.013} & 0.055\scriptsize{$\pm$0.036} & 0.030 & 0.034 & 0.034 & 0.019\scriptsize{$\pm$0.010} & 0.034\scriptsize{$\pm$0.013} & 0.031\scriptsize{$\pm$0.011}					\\
            
            \it{obesity} & 0.024 & 0.013\scriptsize{$\pm$0.009} & 0.184 & 0.064 & 0.069 & 0.094 & 0.114 & 0.136 & 0.239 & 0.035 & 0.163 & 0.150\scriptsize{$\pm$0.125} & 0.362 & 0.181 & 0.184 & 0.047\scriptsize{$\pm$0.008} & {\it\color{blue}0.007}\scriptsize{$\pm$0.006} & {\bf\color{red}0.004}\scriptsize{$\pm$0.004}					\\
            
            \it{cc} & {\bf\color{red}0.001} & 0.004\scriptsize{$\pm$0.003} & 0.005 & 0.012 & 0.010 & 0.016 & 0.006 & 0.007 & 0.005 & 0.004 & {\bf\color{red}0.001} & 0.005\scriptsize{$\pm$0.000} & {\it\color{blue}0.003} & 0.005 & 0.005 & {\bf\color{red}0.001}\scriptsize{$\pm$0.001} & 0.006\scriptsize{$\pm$0.003} & 0.005\scriptsize{$\pm$0.004}					\\
            
            \it{german} & 0.076 & 0.098\scriptsize{$\pm$0.041} & 0.126 & 0.157 & 0.118 & 0.151 & 0.109 & 0.134 & 0.103 & 0.098 & 0.071 & 0.255\scriptsize{$\pm$0.081} & 0.186 & 0.123 & 0.126 & 0.069\scriptsize{$\pm$0.021} & {\it\color{blue}0.068}\scriptsize{$\pm$0.029} & {\bf\color{red}0.067}\scriptsize{$\pm$0.030}					\\
            
            \it{drug} & 0.051 & 0.049\scriptsize{$\pm$0.027} & 0.036 & 0.059 & 0.059 & 0.030 & 0.211 & 0.114 & 0.076 & 0.101 & 0.078 & 0.046\scriptsize{$\pm$0.042} & 0.110 & 0.034 & 0.036 & 0.025\scriptsize{$\pm$0.011} & {\it\color{blue}0.014}\scriptsize{$\pm$0.010} & {\bf\color{red}0.009}\scriptsize{$\pm$0.008}					\\
            
            \it{adult} & 0.151 & 0.059\scriptsize{$\pm$0.003} & 0.103 & {\bf\color{red}0.040} & 0.214 & 0.203 & 0.071 & 0.081 & 0.058 & 0.212 & 0.061 & 0.126\scriptsize{$\pm$0.050} & 0.167 & 0.103 & 0.103 & 0.051\scriptsize{$\pm$0.012} & {\it\color{blue}0.048}\scriptsize{$\pm$0.020} & 0.049\scriptsize{$\pm$0.015}					\\
            
            \it{kdd} & N/A & {\it\color{blue}0.042}\scriptsize{$\pm$0.004} & 0.152 & 0.078 & 0.136 & N/A & 0.052 & 0.243 & 0.047 & 0.146 & 0.152 & 0.102\scriptsize{$\pm$0.061} & 0.087 & 0.152 & 0.152 & 0.015\scriptsize{$\pm$0.003} & 0.054\scriptsize{$\pm$0.005} & {\bf\color{red}0.032}\scriptsize{$\pm$0.009}					\\
            \midrule
            $Score_{F}$ & {\it\color{blue}0.328} & 0.180 & 0.123 & 0.100 & 0.227 & 0.116 & 0.099 & 0.210 & 0.140 & 0.251 & 0.092 & 0.251 & 0.254 & 0.126 & 0.123 & 0.300 & 0.277 & {\bf\color{red}0.402}					 \\
            \bottomrule
        \end{tabular}}\vspace{-2mm}
\end{table*}

We implement our DCFOD in PyTorch\footnote{Source code: https://github.com/brandeis-machine-learning/FairOutlierDetection}. In order to showcase the fairness adaptation effect from the adversarial training, we also compare DCFOD with its independent derivative, DCOD. Other than the absence of subgroup discriminator on DCOD, the two versions share the same implementation. Specifically, we set the number of clusters to 10 and dimension of embedded space to 64. For datasets with sample size greater than 10,000, we set the epoch to 40 and minibatch size to 256 to accelerate training process, otherwise to 90 and 64 respectively. Following~\citet{DEC}, we set the network structure as \textit{drop}-$n$-500-500-2000-64 and 64-2000-500-500-$n$ for encoder and decoder, where \textit{drop} is a dropout layer with rate 0.1. We set the structure for the discriminator to be 64-500-500-2000-$m$, and the cluster centroid layer $\mu$ to be of size (10 * 64). We perform weight initialization on all linear layers with Xavier uniform initialization~\cite{xavier}. Hyperparameter $\alpha$ and $\beta$ are set to 8 and 100 for all datasets. The starting learning rate for encoder, decoder and discriminating layer is set to $1e^{-5}$, that of the centroid layer is set to $1e^{-4}$. Learning rates are multiplied by a factor of 0.1 for every 30 epochs. In the embedded space, we initialize the cluster centroids with MiniBatchKmeans~\cite{mbk} implemented in sklearn. 

During evaluation process, we report the mean and standard deviation for non-deterministic methods by repeating their experiments on 20 random seeds. AE and VAE are not reported as such, since their fluctuations are small enough to be neglected.

\noindent{\textbf{Metrics}}. Since all the outlier detectors return one continuous set of score and all associated datasets provide the ground truth outlier criteria, we use \textit{Area Under Receiver Operating Characteristic Curve (AUC)} to evaluate the outlier diagnostic ability. To measure degree of group fairness, we propose $F_{Gap}$ and $F_{Rank}$, based on the subgroup-wise diagnostic ability gap and the subgroup distribution drift among top-ranked outliers, respectively. Moreover, to thoroughly compare the performance of different methods across datasets, we also propose $Score_{AUC}$ and $Score_{F}$ functions that aggregate the $AUC$ score and fairness measurements on all datasets. 

In specific, \textit{AUC} evaluates the degree of alignment between the outlier score and the ground truth label under varying thresholds:
\begin{equation}
\begin{split}
AUC \coloneqq & 1 - \frac{1}{|\mathcal{T}^+||\mathcal{T}^-|}\sum_{x^+ \in \mathcal{T}^+}\sum_{x^- \in \mathcal{T}^-} \\ & \left(\mathbb{I}(O(x^+) < O(x^-)) + \frac{1}{2}\mathbb{I}(O(x^+) = O(x^-))\right),
\end{split}
\end{equation}
where $\mathcal{T}, \mathcal{T}^+$, and $\mathcal{T}^-$ represent the total sample, true inliers and outliers, respectively, $\mathbb{I}$ is an indicator function and $O(x)$ is the outlier score of data point $x$ given by one outlier detector. In practice, we use $roc\_auc\_score$ in sklearn.metrics\footnote{https://scikit-learn.org} to calculate \textit{AUC}.  

For degree of fairness, $F_{Gap}$ evaluates the diagnostic validity gap between different protected subgroups, where a smaller gap indicates less group bias. Specifically, we calculate the $AUC$ score for each subgroup, and derive $F_{Gap}$ with the difference between the highest $AUC$ and the lowest:
\begin{align}
F_{Gap} \coloneqq \max_{s \in S}  AUC(\mathcal{T}_s) - \min_{s' \in S} AUC(\mathcal{T}_{s'}), 
\end{align}
where $S$ is the sensitive subgroup set, and $\mathcal{T}_s$ represents the sample in subgroup $s$. $AUC(\mathcal{T}_s)$ denotes the $AUC$ score of group $s$.

\begin{figure}[t]
	\centering
	\includegraphics[width=0.425\textwidth]{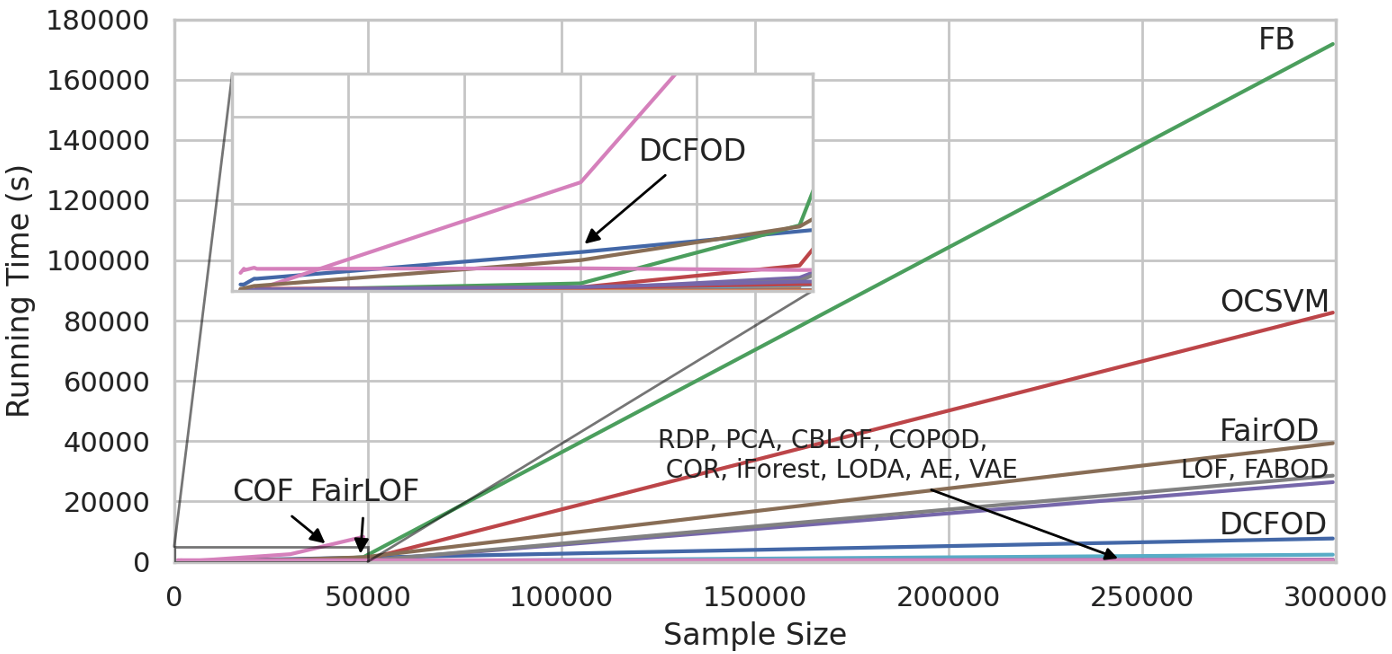}\vspace{-4mm}
	\caption{Execution time comparison on 8 benchmarks.}\label{fig:time}\vspace{-4mm}
\end{figure}

Beyond the performance-related fairness, it is crucial to measure whether sensitive subgroups are distributed fairly among detected outliers. $F_{Rank}$ achieves that goal by calculating the KL-divergence between the subgroup distribution in top outlier candidates in the output and that of the dataset, which is a reference distribution:
\begin{align}
F_{Rank} \coloneqq \max_{r \in R}  \textup{D}_{\textup{KL}}(d_r\Vert \tilde{d}), \quad R = \{5, 6, 7, ..., 20\},
\end{align}
where we rank all output scores from the highest to lowest, with $d_r$ representing the subgroup distribution among the top $r\%$ outlier candidates derived by the outlier detector, and $\tilde{d}$ being the fixed reference distribution of the dataset. Specifically, $F_{Rank}$ records the most significant distribution drift in a range of top $5\%$ to $20\%$ outliers. Both $F_{Gap}$ and $F_{Rank}$ are negative measurements, where the smaller value indicates better fairness degree. 

To better compare model performance under the variability on different datasets, we propose two scoring functions to produce one score for each algorithm on each metric, indicating its aggregated performance across all datasets. We denote the collection of algorithms and datasets with $A, T$, and $A^{(i)}, T^{(j)}$ represent the $i$-th algorithm and $j$-th dataset, respectively. For the diagnostic validity metric $AUC$, we propose $Score_{AUC}$ as the following:
\begin{align}
Score_{AUC}(A^{(i)}) \coloneqq \frac{1}{J}\sum_j \frac{AUC(A^{(i)}, T^{(j)})}{\max_i AUC(A^{(i)}, T^{(j)})},
\end{align}
where $AUC(A^{(i)}, T^{(j)})$ is the $AUC$ of algorithm $A^{(i)}$ on dataset $T^{(j)}$, and $J$ is the total number of datasets. We then propose $Score_{F}$ for the two fairness metrics $F_{Gap}$ and $F_{Rank}$, both denoted as $F$:
\begin{align}
Score_F(A^{(i)}) \coloneqq \frac{1}{J}\sum_j \frac{\min_i F(A^{(i)}, T^{(j)}) + \epsilon}{F(A^{(i)}, T^{(j)}) + \epsilon},
\end{align}
where $\epsilon = 1e^{-5}$. In essence, $Score$ functions rescale numerical results by normalizing each evaluative value using the best result among all algorithms on the same dataset, then average over the scaled value on all datasets. In the end, higher values of $Score_{AUC}$ and $Score_{F}$ indicate better diagnostic validity and more fairness.

\begin{table*}[t]
    \caption{Degree of fairness comparison of 18 algorithms on 8 benchmarks by $F_{Rank}$}\vspace{-4mm}
    \label{tab:rank}
    \centering
    \resizebox{\textwidth}{!}{%
        \begin{tabular}{c|cc|cc|ccc|cc|cccc|ccc|cc}
            \toprule
            \multirow{2}{*}{Dataset} 
                & \multicolumn{2}{c|}{Fair Model} &
                \multicolumn{2}{c|}{Linear Model} & \multicolumn{3}{c|}{Proximity Based} & \multicolumn{2}{c|}{Probability Based} & \multicolumn{4}{c|}{Ensemble Based} & \multicolumn{3}{c|}{Neural Networks} & \multicolumn{2}{c}{Ours}\\ \cmidrule{2 - 19}
            {} & FairLOF & FairOD & PCA & OCSVM & LOF & COF & CBLOF & FABOD & COPOD & FB & iForest & LODA & COR & AE & VAE & RDP  & \textbf{DCOD} & \textbf{DCFOD}\\
            \midrule
            \it{student} & {\bf\color{red}0.003} & 0.016\scriptsize{$\pm$0.010} & 0.009 & 0.067 & {\it\color{blue}0.004} & 0.030 & 0.007 & 0.008 & 0.057 & 0.008 & 0.006 & 0.066\scriptsize{$\pm$0.070} & 0.206 & 0.007 & 0.009 & 0.010\scriptsize{$\pm$0.005} & 0.011\scriptsize{$\pm$0.011} & 0.007\scriptsize{$\pm$0.003}	\\
            
            \it{asd} & 0.012 & 0.030\scriptsize{$\pm$0.026} & 0.023 & 0.013 & 0.013 & {\bf\color{red}0.005} & 0.023 & 0.019 & 0.030 & 0.007 & {\it\color{blue}0.006} & 0.087\scriptsize{$\pm$0.236} & 0.030 & 0.023 & 0.023 & 0.016\scriptsize{$\pm$0.011} & 0.013\scriptsize{$\pm$0.012} & 0.011\scriptsize{$\pm$0.010}		\\
            
            \it{obesity} & 0.031 & 0.008\scriptsize{$\pm$0.007} & 0.013 & 0.030 & 0.015 & 0.018 & 0.073 & 0.009 & 0.067 & 0.009 & 0.011 & 0.057\scriptsize{$\pm$0.187} & 0.669 & 0.012 & 0.013 & {\bf\color{red}0.003}\scriptsize{$\pm$0.001} & 0.008\scriptsize{$\pm$0.007} & {\it\color{blue}0.005}\scriptsize{$\pm$0.003}\\ 
            
            \it{cc} & {\bf\color{red}0.001} & {\bf\color{red}0.001}\scriptsize{$\pm$0.000} & 0.010 & 0.010 & 0.004 & 0.003 & 0.014 & 0.012 & 0.026 & {\it\color{blue}0.002} & 0.024 & 0.049\scriptsize{$\pm$0.001} & 0.112 & 0.010 & 0.010 & {\it\color{blue}0.002}\scriptsize{$\pm$0.000} & {\bf\color{red}0.001}\scriptsize{$\pm$0.000} & {\bf\color{red}0.001}\scriptsize{$\pm$0.000}\\
            
            \it{german} & 0.101 & 0.038\scriptsize{$\pm$0.017} & {\bf\color{red}0.014} & 0.105 & 0.102 & 0.046 & 0.061 & 0.054 & 0.024 & 0.148 & 0.042 & 0.029\scriptsize{$\pm$0.560} & 0.255 & {\bf\color{red}0.014} & {\it\color{blue}0.015} & 0.016\scriptsize{$\pm$0.002} & 0.048\scriptsize{$\pm$0.018} & 0.044\scriptsize{$\pm$0.022}					\\
            
            \it{drug} & 0.003 & 0.007\scriptsize{$\pm$0.006} & 0.003 & 0.009 & 0.010 & {\bf\color{red}0.001} & {\it\color{blue}0.002} & 0.005 & 0.009 & 0.004 & 0.009 & 0.093\scriptsize{$\pm$0.324} & 0.091 & 0.003 & 0.003 & 0.005\scriptsize{$\pm$0.002} & 0.008\scriptsize{$\pm$0.006} & 0.008\scriptsize{$\pm$0.008}					\\
            
            \it{adult} & 0.131 & {\bf\color{red}0.001}\scriptsize{$\pm$0.000} & 0.483 & {\it\color{blue}0.009} & 0.154 & 0.153 & 0.071 & 0.244 & 0.662 & 0.045 & 0.562 & 0.073\scriptsize{$\pm$0.344} & 0.014 & 0.483 & 0.483 & {\bf\color{red}0.001}\scriptsize{$\pm$0.000} & {\bf\color{red}0.001}\scriptsize{$\pm$0.000} & {\bf\color{red}0.001}\scriptsize{$\pm$0.000}\\
            
            \it{kdd} & N/A & {\it\color{blue}0.001}\scriptsize{$\pm$0.000} & 0.185 & 0.022 & 0.023 & N/A & 0.016 & 0.003 & 0.235 & 0.057 & 0.140 & 0.095\scriptsize{$\pm$0.112} & 0.047 & 0.185 & 0.185 & {\bf\color{red}0.000}\scriptsize{$\pm$0.000} & {\bf\color{red}0.000}\scriptsize{$\pm$0.000} & {\bf\color{red}0.000}\scriptsize{$\pm$0.000}					\\
            \midrule
            $Score_{F}$ & 0.370 & 0.236 & 0.279 & 0.107 & 0.215 & {\it\color{blue}0.391} & 0.179 & 0.191 & 0.126 & 0.246 & 0.084 & 0.246 & 0.276 & 0.292 & 0.279 & 0.325 & 0.346 & {\bf\color{red}0.442}			\\
            \bottomrule
        \end{tabular}}\vspace{-4mm}
\end{table*}

\noindent\textbf{Environment}. Experiments were run on a 12 core processor Linux server with Intel Core i7-6850K@3.60 Ghz and 64GB RAM.

\begin{figure*}[t]
  \centering
    \subfigure[\textit{asd}]{
    \includegraphics[width=0.495\textwidth, height = 3.2cm]{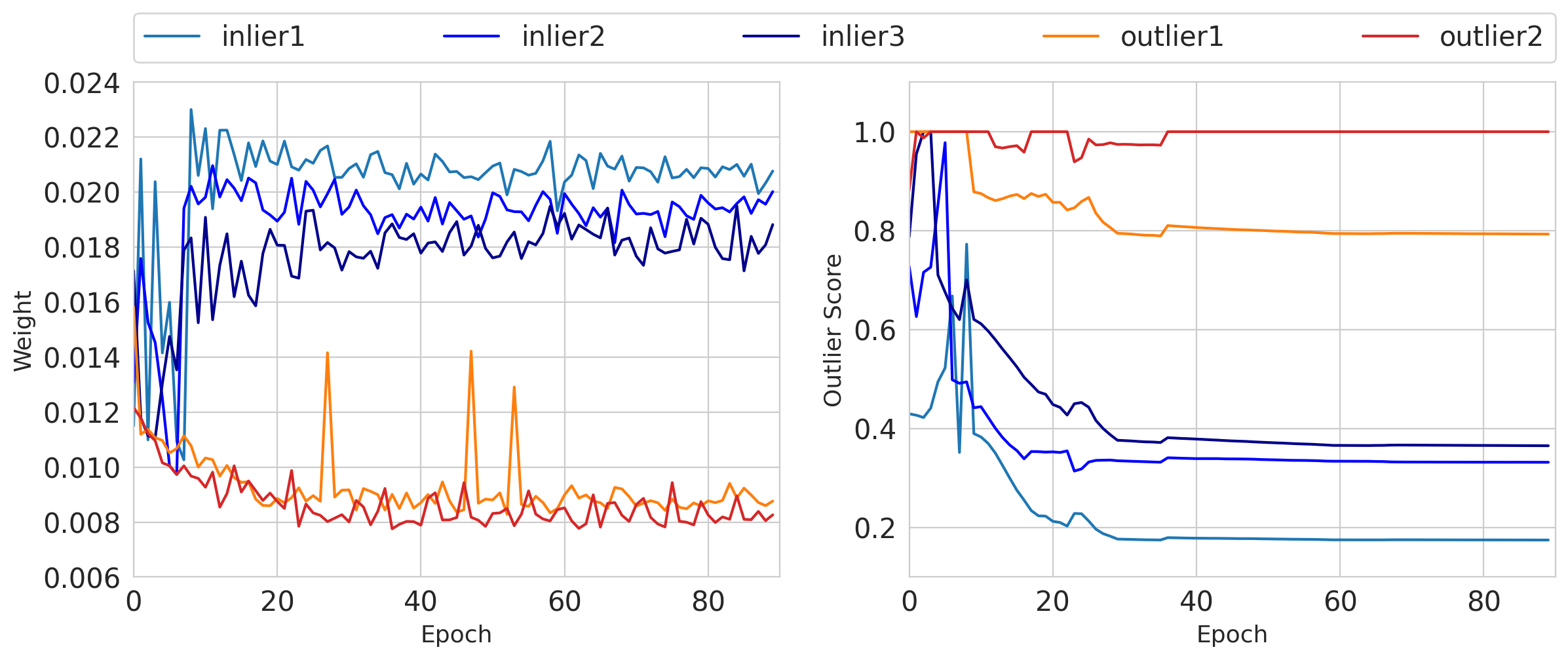}\label{fig:asdweight}}\hspace{-1.5mm}
    \subfigure[\textit{german}]{
    \includegraphics[width=0.495\textwidth, height = 3.2cm]{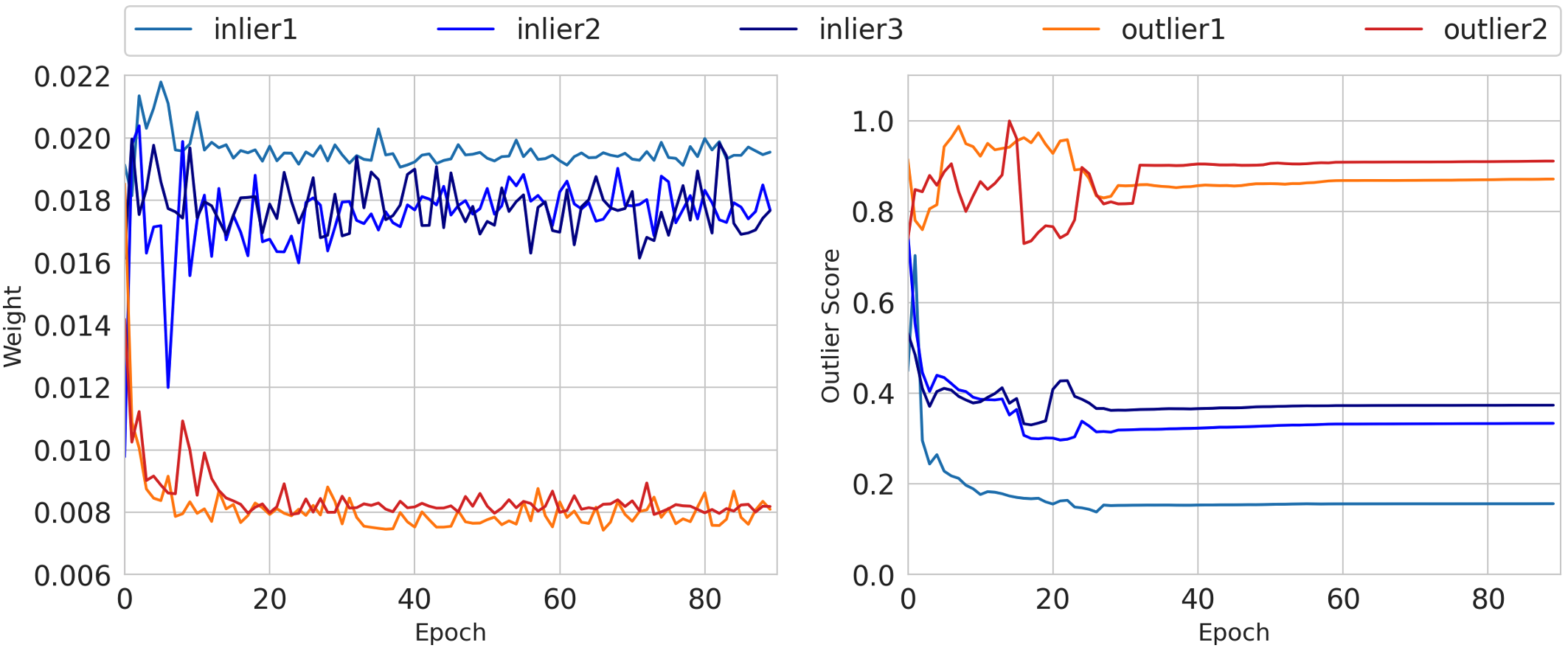}\label{fig:germanscore}}\vspace{-4mm}
  \caption{Weight and score trends of inliers and outliers during iterations on \textit{asd} and \textit{german}.} \label{fig:weight_score}\vspace{-4mm}
\end{figure*}

\subsection{Performance}
Table~\ref{tab:auc}, \ref{tab:gap} and \ref{tab:rank} show the performance comparison of 18 algorithms on 8 benchmarks in terms of $AUC$, $F_{Gap}$ and $F_{Rank}$, where the best and second best results are highlighted in red bold and blue italic. 

For outlier validation validity, different algorithms achieve the best results on different datasets. OCSVM gets 0.621 on \textit{cc}, LOF and FABOD achieve 0.589 on \textit{german}. Although these algorithms are hardly considered state-of-the-art, they are still effective in some scenarios. FairOD gets a 0.605 on \textit{adult}, which shows its fairness adaptation can also bring high detection accuracy in certain cases. Our baseline method DCOD delivers two best results out of eight datasets, while our proposed DCFOD provides three best results and excels other competitive methods by a large margin on the comprehensive measurement $Score_{AUC}$. The demonstrated superior performance on DCFOD comes from two main aspects: (1) our model adopts a well-generalized clustering based framework, which holds a strong coupling with the unsupervised outlier detection task. (2) The weighted representation learning is capable of generating a feature embedding that captures the unique intrinsic cluster structure for each dataset. In particular, the dynamic weight module mitigates the influences from outliers' abnormal representations and fosters a task-favorable feature space. 

For fairness measurements, DCFOD has the highest $Score_F$ values on both $F_{Gap}$ and $F_{Rank}$. Specifically, our model delivers the best results on \textit{obesity}, \textit{german}, \textit{drug} and \textit{kdd} by $F_{Gap}$, where the adversarial training shrinks the detection validity gap between subgroups by an average of 30$\%$ compared to DCOD. Other competitive methods, including FairLOF, OCVSM, LOF, FB, iForest, and RDP, share some best results on other datasets. Similar phenomena occurs on $F_{Rank}$, as shown in Table~\ref{tab:rank}, where DCFOD achieves the most fair performance on \textit{cc}, \textit{adult}, and \textit{kdd}. On the rest datasets, the adversarial training also generates a fairer subgroup distribution compared to DCOD and improves $F_{Rank}$ by an average of $10\%$.
\begin{figure}[t]
	\centering
	\subfigure[\textit{AUC}]{
	\includegraphics[width=0.232\textwidth, height = 3cm]{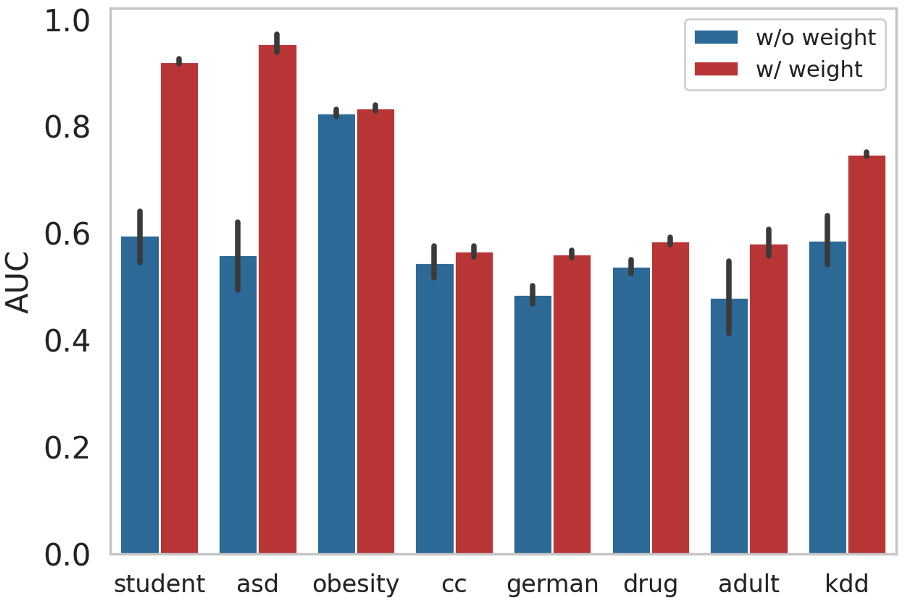}\label{fig:weight_auc}}\hspace{-1mm}
	\subfigure[$F_{Gap}$]{
	\includegraphics[width=0.232\textwidth, height = 3cm]{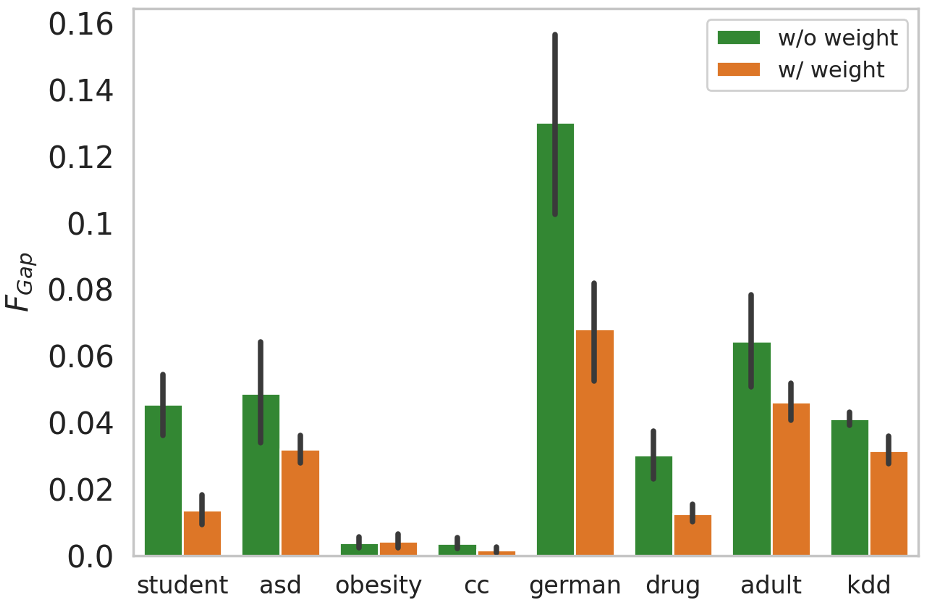}\label{fig:weight_gap}}\vspace{-4mm}
	\caption{Effectiveness of weight adjustment on \textit{AUC} and fairness measurement $F_{Gap}$. }\label{fig:weight}\vspace{-5.5mm}
\end{figure}

\begin{figure*}[t]
  \centering
    \subfigure[\textit{student}]{
    \includegraphics[width=0.246\textwidth, height = 3.2cm]{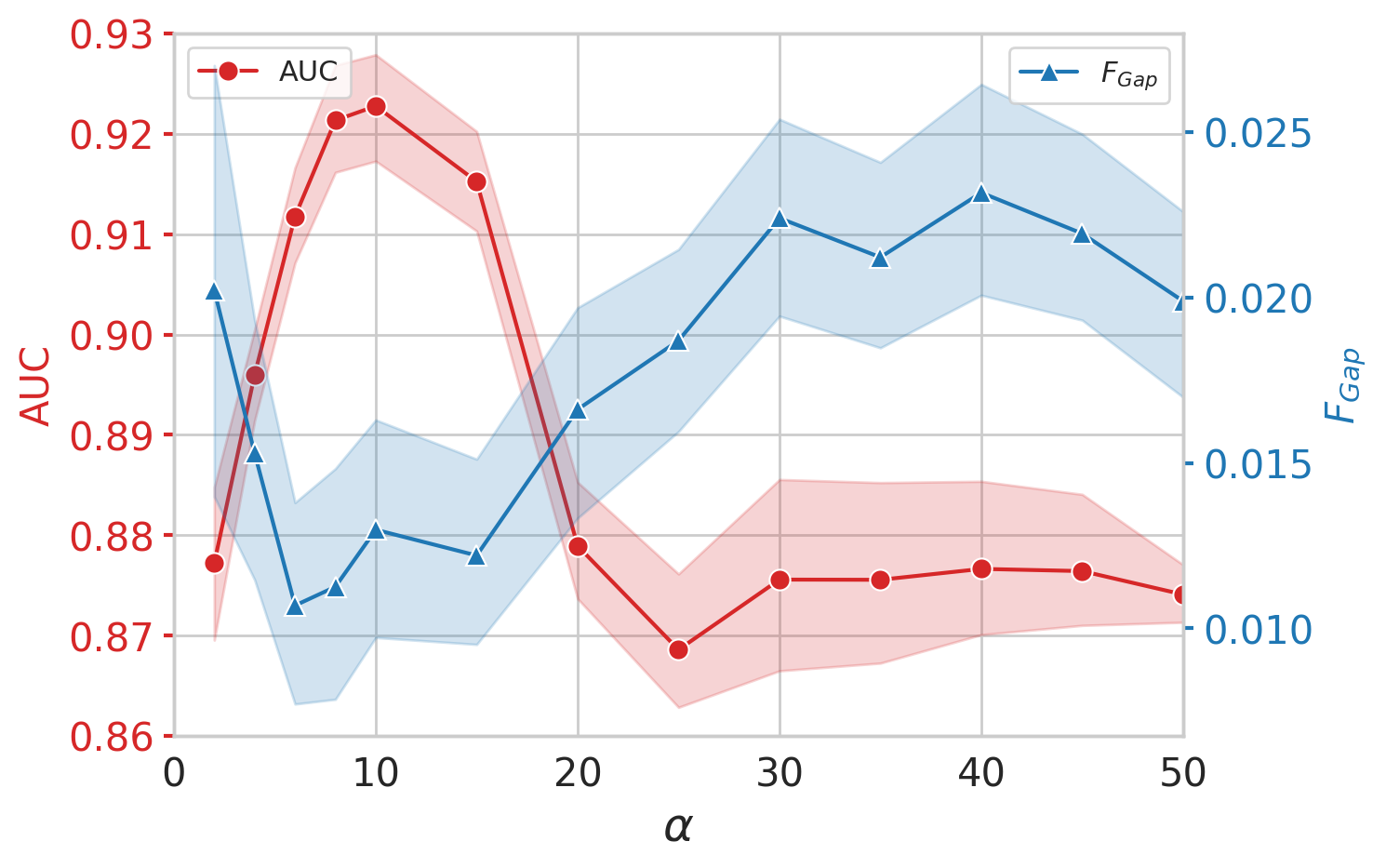}\label{fig:hypers_1}}\hspace{-1.5mm}
    \subfigure[\textit{asd}]{
    \includegraphics[width=0.246\textwidth, height = 3.2cm]{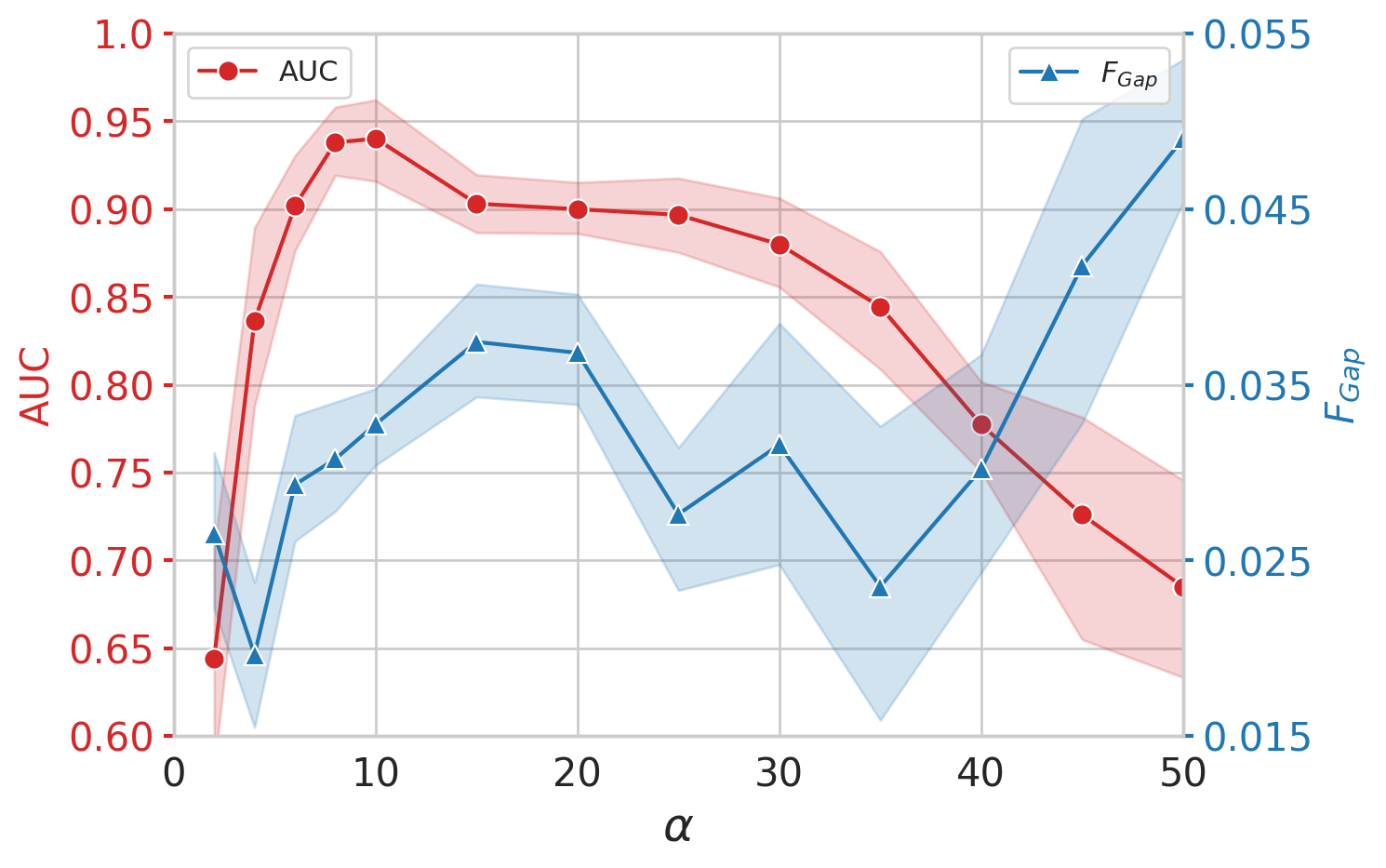}\label{fig:hypers_2}}\hspace{-1.5mm}
    \subfigure[\textit{obesity}]{
    \includegraphics[width=0.245\textwidth, height = 3.2cm]{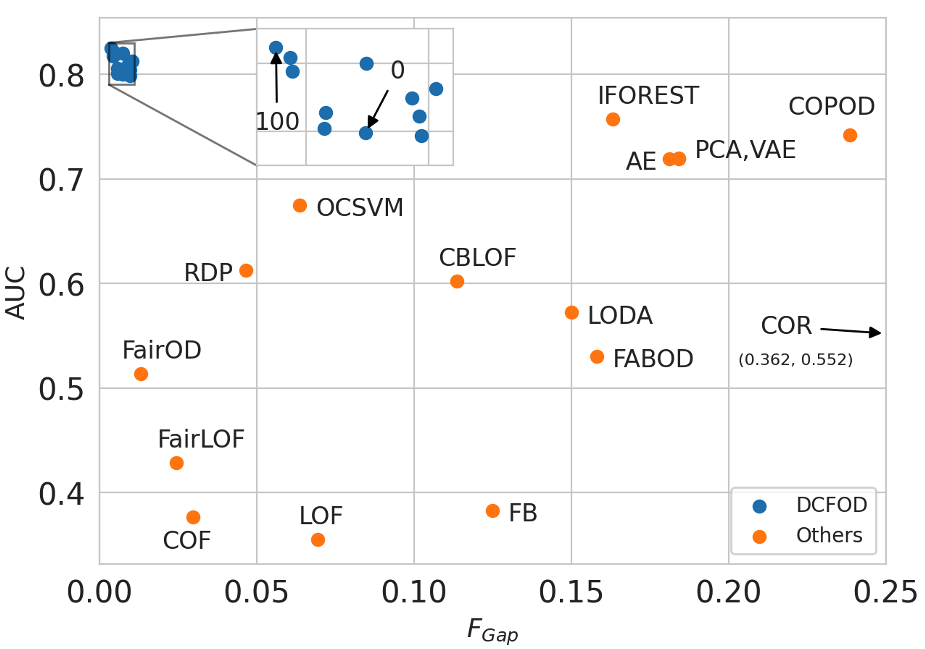}\label{fig:hyperf_1}}\hspace{-1.5mm}
    \subfigure[\textit{drug}]{
    \includegraphics[width=0.247\textwidth, height = 3.2cm]{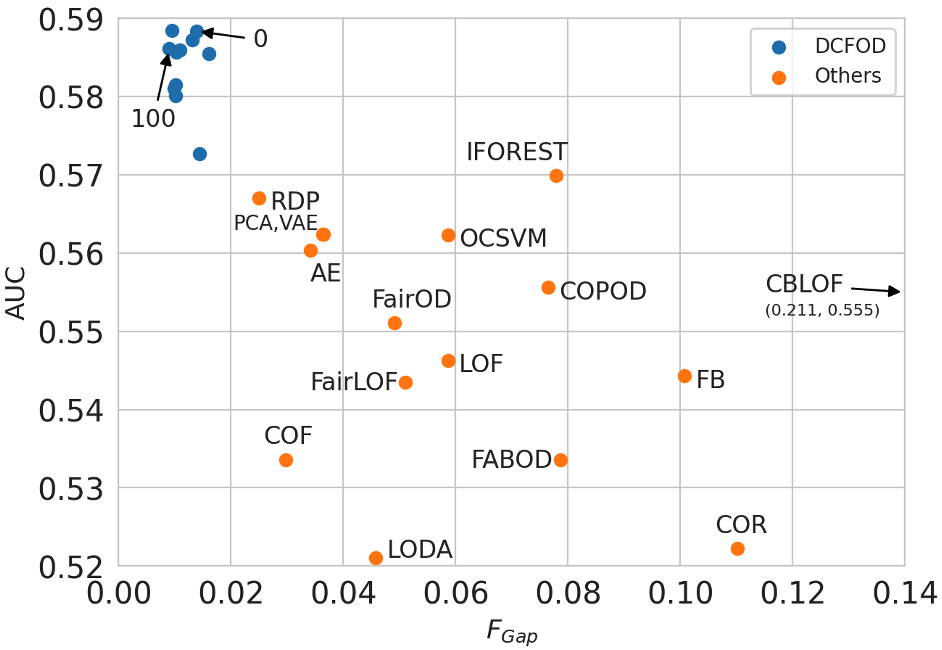}\label{fig:hyperf_2}}\vspace{-4mm}
  \caption{Hyperparameter analysis for $\alpha$ and $\beta$. (a) \& (b) The trends of \textit{AUC} and $F_{Gap}$ on \textit{student} and \textit{asd}, while $\alpha$ ranging from [2, 4, 6, 8, 10, 15, 20, 25, 30, 35, 40, 45, 50] and $\beta$ pinned at 100. (c) \& (d) Pareto diagram of \textit{AUC} and $F_{Gap}$ for all competitive methods on \textit{obesity} and \textit{drug}. $\beta$ ranges from [0, 1, 10, 20, 50, 80, 100, 150, 200, 250, 300], while $\alpha$ pinned at 8. Points with $\beta = 0$ and $\beta = 100$ are annotated with arrows, which are the model w/o adversarial training and the pareto front, respectively.} \label{fig:hyper}\vspace{-4mm}
\end{figure*}

We then discuss FairLOF and FairOD, two recently proposed fair outlier detectors. They set out to sacrifice certain degrees of detection validity in exchange for more fairness compared to their base methods LOF and autoencoder, which has been demonstrated in our experiment. Compared to LOF, FairLOF has a lower \textit{AUC} with improved degrees of fairness on both $F_{Gap}$ and $F_{Rank}$, while FairOD has a slightly lower \textit{AUC} score compared to that of AE, with a similar $F_{Rank}$ and improved $F_{Gap}$. The above models both apply a post-processing technique that seeks a consistent outlier ranking with a baseline detector, whose capacity might restrict the fair model from pursuing a higher detection accuracy. In contrast, DCFOD adopts a baseline-free, in-processing procedure to learn a fair feature representation where the detection accuracy is simultaneously optimizable. As shown in Table~\ref{tab:auc}, the fairness-adapted DCFOD surpasses DCOD on $AUC$, which demonstrates that the goals of accurate diagnosis and fairness adaptation may be beneficial to each other in a proper feature space. Based on the above illustrations, with the representation learning framework and fairness-adversarial training, the legitimacy of DCFOD's superior performance on both the detection validity and group fairness is well demonstrated. 

We compare the execution time of all methods in Figure~\ref{fig:time}. On datasets with less than 50,000 instances, all algorithms except COF have fast performances. Several maintain a low running time across all datasets, shown at the bottom lines in the figure. As sample size increases, FairLOF and COF encounter an out-of-memory error, and several other methods dramatically increase their running time. Meanwhile, DCFOD remains stable and keeps a flat increase rate due to its linear time complexity to the sample size $N$ and parameters $D, K$.

\subsection{Factor Exploration}
We hereby explore the impact of critical factors in DCFOD. We look into the weight module for its quantitative effects on model training and performance, then test model volatility on hyperparameters. We use $AUC$ and $F_{Gap}$ performance for visualization due to the space limitation.

\noindent\textbf{Weight and Outlier Score}. The proposed DCFOD model assigns a dynamic weight for each instance during training. The likely-outlier candidates that are far away from the cluster centroids have higher outlier scores while assigned with lower weights. With this approach, we alleviate outliers' disturbance toward the overall cluster structure and improve the model performance. Figure~\ref{fig:weight_score} shows the dynamic relationship between outlier scores and weight adjustments on \textit{asd} and \textit{german}. For each dataset, we choose three inliers and two outliers for demonstration, where consistent trends are exhibited. After several iterations, the weights and scores stabilize into two distinct groups. For outliers, scores converge to high values, and weights go to the lowest value of 0.008, while inliers have more dominant weights at around 0.02 and low outlier scores. The demonstration uses a minibatch of size 64, where one data point's original contribution is $1/64 \approx 0.016$. Our adaptive weight module reinforces inliers' contributions by approximately $25\%$ and decreases that of outliers by $50\%$, respectively.

Figure~\ref{fig:weight} compares \textit{AUC} and $F_{Gap}$ on all datasets with respect to the weight module's presence, whose beneficial effects towards both the detection validity and fairness degree are clearly exhibited. Particularly, the weight module brings an average of $46\%$ increase in \textit{AUC} on \textit{student}, \textit{asd}, and \textit{kdd} datasets, while decreasing the detection validity gap between subgroups by an average of $55\%$ on \textit{student}, \textit{german}, and \textit{drug}. The weight module's enhancement in outlier validation invariably applies to all subgroups, thus closing their validity gaps and making the model fairer. 
\begin{figure}[t]
	\centering
	\subfigure[\textit{AUC}]{
	\includegraphics[width=0.232\textwidth, height = 3cm]{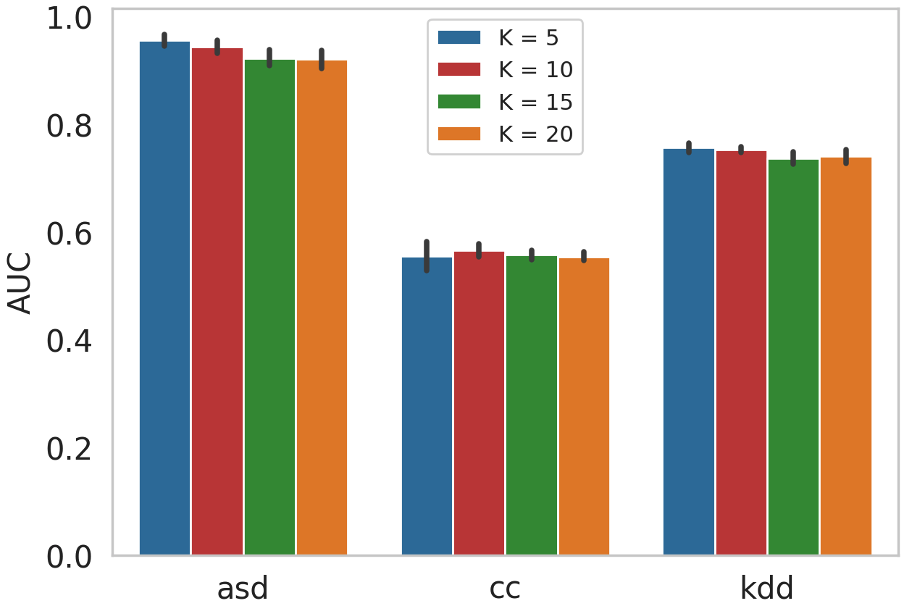}\label{fig:cluster_auc}}\hspace{-1mm}
	\subfigure[$F_{Gap}$]{
	\includegraphics[width=0.232\textwidth, height = 3cm]{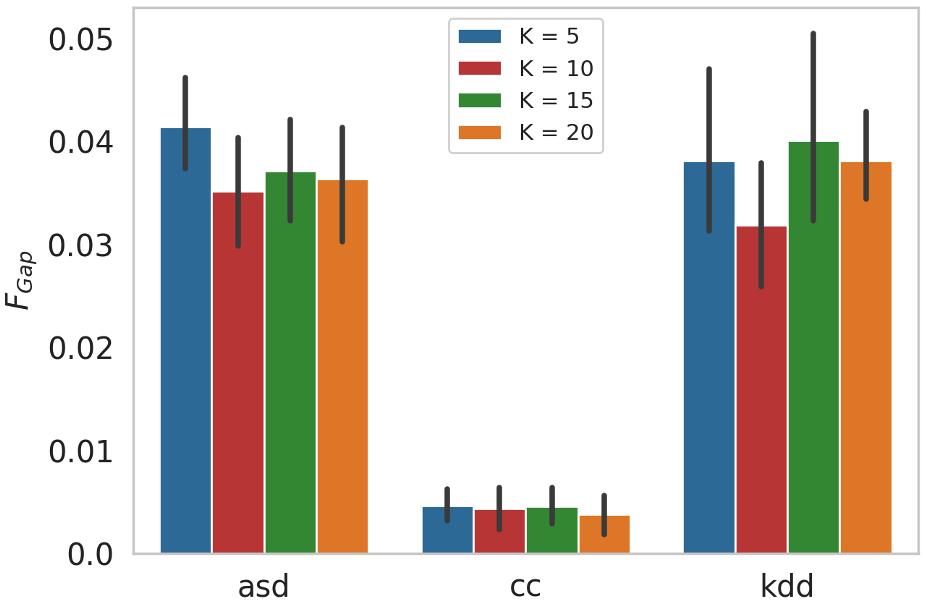}\label{fig:cluster_gap}}\vspace{-4mm}
	\caption{Performance of DCFOD with different cluster numbers on \textit{asd}, \textit{cc} and \textit{kdd} by $AUC$ and $F_{Gap}$.}\label{fig:cluster}\vspace{-6mm}
\end{figure}

\noindent\textbf{Hyperparameter Analysis}. In our objective function, $\alpha$ and $\beta$ govern the balance between the self-reconstruction loss and fairness-adversarial loss. We hereby explore the volatility of the model with respect to changes on hyperparameters and verify the legitimacy of the default values (i.e., $\alpha = 8, \beta = 100$). As shown in Figure~\ref{fig:hypers_1} and ~\ref{fig:hypers_2}, the model reaches the best combination between $AUC$ and $F_{Gap}$ when $\alpha = 8$. As $\alpha$ further increases, detection validity decreases and the subgroup validity gap subsequently enlarges. Figure~\ref{fig:hyperf_1} and ~\ref{fig:hyperf_2} are the pareto diagrams of $AUC$ and $F_{Gap}$ on dataset \textit{obesity} and \textit{drug}. DCFOD is observed to consistently yield results that are both fair and of high detection validity under varying $\beta$, compared to all other competitive methods. $\beta = 100$ takes the pareto front on both datasets, with clear improvements on fairness degree $F_{Gap}$ compared to $\beta = 0$ (w/o adversarial training).

Another hyperparameter in our model is the number of clusters $K$. In a typical Clustering based algorithm, it is set as the number of classes. However, in real-world unsupervised outlier detection tasks, we are often not provided with the actual number of classes. In our experiment, we uniformly set the number of clusters $K=10$ for all datasets. Figure~\ref{fig:cluster} shows the $AUC$ and $F_{Gap}$ on \textit{cc}, \textit{kdd}, and \textit{asd} with different cluster numbers. The consistent performance shows that our DCFOD is not sensitive to the cluster number. 

\section{Conclusion}
We studied the fairness issues in the task of unsupervised outlier detection. We proposed a Deep Clustering based Fair Outlier Detection framework that conducts downstream task-favorable representation learning with adversarial training to optimize the detection validity concurrently with group fairness. Specifically, we designed an adaptive weight module to administer model training by reinforcing the likely-inliers’ contributions while alleviating the outliers’ negative effects. We extended the existing fairness evaluative criteria in the context of outlier detection and proposed two metrics regarding group-wise diagnostic accuracy gap and subgroup distribution among likely-outliers. Experiments showcased our model’s superiority over the recent fair models and conventional outlier detection methods in both outlier validation and degree of fairness.

\section{Acknowledgment}
This work was supported in part by NSF OAC 1920147.

\bibliographystyle{ACM-Reference-Format}
\bibliography{ref}

\end{document}